\documentclass[sigconf]{acmart}
\AtBeginDocument{%
  }

\setcopyright{acmlicensed}
\copyrightyear{2025}
\acmYear{2025}
\acmDOI{XXXXXXX.XXXXXXX}

\acmConference[Conference acronym 'XX]{Make sure to enter the correct
  conference title from your rights confirmation emai}{June 03--05,
  2018}{Woodstock, NY}
\acmISBN{978-1-4503-XXXX-X/18/06}

\begin{document}

\title{RACCOON: A Retrieval-Augmented Generation Approach for Location Coordinate Capture from News Articles}

\author{Jonathan Lin, Aditya Joshi, Hye-young Paik, Tri Dung Doung, Deepti Gurdasani}
\affiliation{%
  \institution{University of New South Wales}
  \city{Sydney}
  \country{Australia}}
\email{jonathan.lin@student.unsw.edu.au, aditya.joshi@unsw.edu.au, h.paik@unsw.edu.au}
\email{duong.dung@unsw.edu.au, dgurdasani@gmail.com}

\renewcommand{\shortauthors}{Lin et al.}

\begin{abstract}
Geocoding involves automatic extraction of location coordinates of incidents reported in news articles, and can be used for epidemic intelligence or disaster management. This paper introduces Retrieval-Augmented Coordinate Capture Of Online News articles (RACCOON), an open-source\footnote{https://github.com/jonlin223/RACCOON-geocoding} geocoding approach that extracts geolocations from news articles. RACCOON uses a retrieval-augmented generation (RAG) approach where candidate locations and associated information are retrieved in the form of context from a location database, and a prompt containing the retrieved context, location mentions and news articles is fed to an LLM to generate the location coordinates.  Our evaluation on three datasets, two underlying LLMs, three baselines and several ablation tests based on the components of RACCOON demonstrate the utility of RACCOON. To the best of our knowledge, RACCOON is the first RAG-based approach for geocoding using pre-trained LLMs.
\end{abstract}

\ccsdesc[500]{Computing methodologies~Natural language processing}
\ccsdesc[500]{Computing methodologies~Language resources}
\ccsdesc[500]{Computing methodologies~Information extraction}
\ccsdesc[300]{Applied computing~Health informatics}

\keywords{RAG, large language models, geocoding, location extraction, news articles}

\received{20 February 2007}
\received[revised]{12 March 2009}
\received[accepted]{5 June 2009}

\maketitle

\section{Introduction}
Online content such as news articles and social media posts have been used for applications such as epidemic intelligence~\cite{joshi2019survey}. Identifying geographical coordinates corresponding to locations mentioned in the news articles is important for localising incident reports and understanding affected areas for these applications. Geocoding, also known as toponym resolution, is the task of extracting geo-coordinates of location references mentioned in a text\footnote{We use the term `geocoding' in the rest of the paper.}. Geocoding is challenging due to ambiguity of locations. Location references may share names with non-locations (`Washington' can be either a person or place), distinct locations may share names (`Newcastle' in the UK and Australia), or locations may have alternate names or abbreviations (`Royal Melbourne Hospital' or `RMH' are the same location). Modeled as a typical information extraction task, geocoding has been performed using rule-based, classical machine learning and deep learning-based approaches, as described in the survey by ~\citet{zhang2024survey}. However, retrieval-augmented generation (RAG)~\cite{lewis2020retrieval} has been found to improve the accuracy of responses generated by LLMs. Therefore, this paper introduces an approach for geocoding, named as Retrieval-Augmented Coordinate Capture Of Online News articles (RACCOON).  The novelty of RACCOON lies in the use of RAG, coupled with techniques such as population re-ranking and candidate retrieval, inspired from past work in geocoding. RACCOON is evaluated on three benchmark datasets, and compared against three baselines (including a rule-based and an LLM prompting-based approach where the latter does not use RAG) via a suite of metrics. Ablation tests on RACCOON include testing alternate LLMs (namely, GPT and Gemini) and the role of other components of RACCOON.

\begin{figure*}
    \centering
    \includegraphics[width=0.75\linewidth]{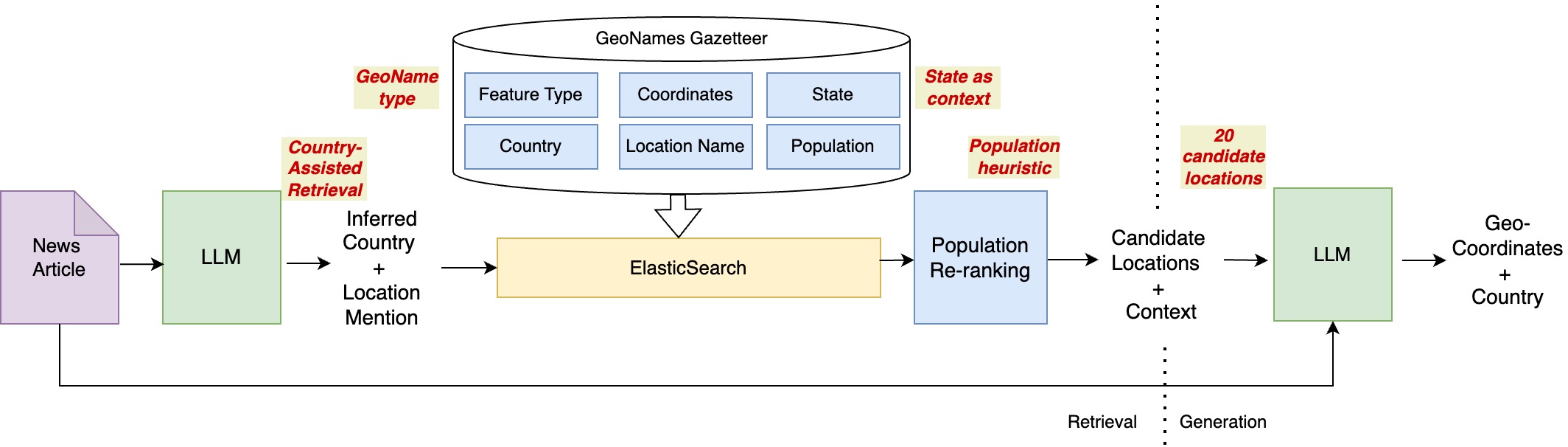}
    \caption{Architecture of RACCOON.}
    \label{fig:rag-final}
\end{figure*}
\section{Related Work}
\citet{zhang2024survey} is a survey that decomposes toponym resolution into two categories: ranking and classification approaches. Ranking-based approaches query geospatial databases (gazetteers) and re-rank entries based on specific features, However, these methods often encounter challenges such as limited retriever recall and gazetteer incompleteness \cite{grover2010use, wang2019dm_nlp, halterman2023mordecai}. Recent advancements such as GeoNorm \cite{zhang2023improving} leverage Transformer-based architectures and employ a two-step retrieval-then-reranking framework. RACCOON further integrates RAG to enhance the performance on geocoding. However, retrieval-based models like GeoNorm face challenges such as retriever recall and gazetteer incompleteness, which RACCOON addresses by incorporating GeoNorm's two-step method into a Retrieval Augmented Generation (RAG) framework for LLMs. Similarly, GeoNorm involves a train-test setup while RACCOON uses an off-the-shelf pretrained LLM. In contrast, classification approaches make geospatial predictions when the Earth's surface has been discretised into tiny areas. Geocoding is then models as a classification task with the output being one of these tiles \cite{delozier2015gazetteer}. While classification approaches do not encounter gazetteer-related issues, they cannot leverage direct geospatial information from gazetteers. By leveraging the pre-trained geospatial knowledge present in LLMs, RACCOON can similarly avoid these gazetteer-related issues. In contrast, ranking-based approaches query a reference geospatial database (gazetteers) and re-rank the entries based on features. Finally, previous studies in geocoding include rule-based \cite{grover2010use}, machine-learning \cite{wang2019dm_nlp}, and deep-learning \cite{halterman2023mordecai} approaches, with recent models like GeoNorm \cite{zhang2023improving} as described above.


\section{Methodology}
RACCOON is implemented using LangChain, and the architecture is shown in Figure \ref{fig:rag-final}. RACCOON uses RAG to connect the GeoNames\footnote{https://www.geonames.org/; Accessed on 5th December 2024.} gazetteer to an LLM, providing it with specific geospatial knowledge that it might not be able to retrieve when only using an LLM via prompts. 
RACCOON consists of five key components: (a) country-assisted retrieval,  (b) 20 candidate location entries, (c) GeoNames feature types, (d) population heuristic, and (e) state as a context. They are highlighted in Figure \ref{fig:rag-final}, and also indicated in boldface in the subsequent description\footnote{Note that, our results show ablation tests on these five components (along with choice of the LLM as a sixth) to validate their value.}.

\paragraph{Retriever} Given a news article as input, a GPT-4o-mini model is used to identify a country for each location mention (\textbf{country-assisted retrieval}). The inferred country is passed along with the location mention, and helps to narrow down the candidate locations, as recommended in \citet{zhang2023improving}. The locations are then searched in a Lucene search engine, based on the BM-25 ranking function. ElasticSearch index forms the background information for the retriever.  The index contains location names, coordinates, country and population, English language alternate names, and their abbreviations.  RACCOON sorts the retrieved matches using the \textbf{population heuristic} (i.e., based on population), taking \textbf{20 top entries} as candidate locations.

\paragraph{Generation} Once the candidate locations have been used, RACCOON uses in-context prompting of the GPT-4o-mini model. The prompt consists of three parts: (a) A sentence containing identifying information about the candidate location such as name, coordinates, along with additional disambiguating features such as \textbf{GeoNames feature type}, country and \textbf{state}; (B) Location reference; and (C) News article. Specific prompt text can be found in the github repository linked in the abstract.  Note that we include the annotated location references from the evaluation datasets in the prompt, assuming that location references are available. This is not unrealistic since NER models can be used to extract location references. 

\begin{table}[]
\begin{tabular}[width=\textwidth]{p{2.5cm}p{1.5cm}p{2cm}p{1.5cm}}
\toprule
                                      & \textbf{GeoVirus} & \textbf{GeoWebNews} & \textbf{LGL} \\ \midrule
\# Articles                    & 229               & 200                 & 588          \\ \hline
\# Toponyms         & 2167              & 1447                & 4463         \\ \hline
\# Unique Considered Locations & 685               & 666                 & 1086         \\ \hline
Median Population of Locations\footnote{Population figures taken from GeoNames. Locations with no GeoNames annotation excluded. GeoVirus excluded as it is not annotated with GeoNames IDs.}        & N/A               & 3 522 465           & 206 922      \\ \bottomrule
\end{tabular}%
\caption{Dataset statistics.}
\label{tab:dataset-comp}
\end{table}
\section{Experiment Setup}
\paragraph{Datasets}
We evaluated RACCOON on three datasets (statistics in Table \ref{tab:dataset-comp}): GeoVirus \citep{gritta2018melbourne} GeoWebNews \citep{gritta2018pragmatic} and Local-Global Lexicon (LGL) \citep{lieberman2010geotagging}. These datasets contain English news articles that are annotated with location references and associated coordinates for each reference. For each location reference, we parse location name, latitude, and longitude within the news text. For GeoWebNews and LGL, we also parse each location's GeoNames ID. We then use this ID and the resultant gazetteer entry to include the country in which the location is contained. As GeoVirus is not annotated with GeoNames IDs, we include country information through reverse geocoding with the Nominatim geocoder. Since GeoWebNews annotates many toponyms, including those that do not refer to locational concepts, we only parse those toponyms which have type `Literal', `Literal Modifier', `Mixed', `Coercion', and `Embedded Literal' since they refer to locational concepts \cite{gritta2018pragmatic}. For LGL, we exclude any toponyms which are not labeled with associated coordinates. We did not exclude any entries from GeoVirus.

\paragraph{Metrics} Our evaluation metrics are based on error. Error is defined as the distance between a location as defined by the evaluation dataset and as inferred by a geocoding model. We use a combination of metrics as typically used for geocoding\cite{zhang2024survey}.  (A) \textbf{Mean error} is the average error across all the news articles in the test set; (B) \textbf{Accuracy @161km} is the percentage of errors that are underneath 161km; (C) \textbf{Country Accuracy} is the percentage of locations that have been resolved to the correct country. If a location is not contained within a single country such as a continent, region or ocean, we do not include it into this measure. (D) \textbf{Area Under the Curve (AUC)} is a measure commonly used in geocoding studies. Unlike other metrics, AUC applies the logarithmic function to errors.

\[\ AUC={{\int_{0}^{dim(x)}\ln{(x)}dx}\over{dim(x)}\times \ln{(20039)}}\]

In addition, we also report (E) the number of returned responses as a recall-oriented metric, since the models (the baselines in particular) return empty responses for a significantly higher number of test instances.


\paragraph{Baselines}
\begin{table*}
\begin{tabular}{p{2.4cm}p{0.6cm}p{0.6cm}p{0.6cm}p{0.6cm}p{0.6cm}p{0.6cm}p{0.6cm}p{0.6cm}p{0.6cm}p{0.6cm}p{0.6cm}p{0.6cm}p{0.6cm}p{0.6cm}p{0.6cm}p{0.6cm}}
\toprule
\multicolumn{1}{c}{}      & \multicolumn{5}{c}{GeoVirus}                                               & \multicolumn{5}{c}{GeoWebNews}                                           & \multicolumn{5}{c}{LGL}                                                   \\ 
\multicolumn{1}{c}{Model} & MErr           & A@161      & CAcc         & AUC            & Num      & MErr            & A@161          & CAcc           & AUC            & Num   & MErr            & A@161          & CAcc           & AUC            & Num  \\ \midrule
\multicolumn{15}{c}{\bf Baselines}\\ \midrule
Gazetteer Base      & 919.89          & 0.697          & 0.889          & 0.402          & 1885 & 873.69          & 0.721          & 0.891          & 0.3411         & 1028                   & 1168.79         & 0.633          & 0.859          & 0.381          & 3050 \\
LLM Base            & 159.34          & 0.792          & 0.969 & 0.372          & 1668  & 107.91 & 0.871          & \textbf{0.998} & 0.248          & 892   & \textbf{203.86} & \textbf{0.901} & 0.976          & 0.224          & 2964 \\
RAG Base            & 162.79          & 0.839          & 0.957          & 0.293          & 2165   & 342.34          & 0.885          & 0.971          & 0.133          & 1443     & 472.55          & 0.831          & 0.951          & 0.203          & 4459 \\ \midrule
\multicolumn{15}{c}{\bf Our Approach}\\ \midrule
RACCOON           & \textbf{124.19} & \textbf{0.861} & 0.958          & \textbf{0.270} & 2166   & 207.76          & .903 & 0.986          & 0.119 & 1439  & 284.10          & 0.849          & 0.979 & 0.175 & 4439 \\ \midrule
\multicolumn{15}{c}{\bf Ablations of Our Approach}\\ \midrule
( - 20 -> 1 Candidate Entry)          & 149.80          & 0.808          & 0.957          & 0.306          & 2166   & 266.85          & 0.856          & 0.984          & 0.168          & 1442       & 360.38          & 0.796          & 0.977          & 0.239          & 4444 \\
(- Country-Assisted Retrieval) & 146.09          & 0.857          & 0.955          & 0.278          & 2167       & 269.96          & 0.897          & 0.976          & 0.126          & 1440   & 524.94          & 0.822          & 0.946          & 0.202          & 4445 \\
(- GeoNames Feature Types)     & 131.99          & 0.858          & 0.961          & 0.274          & 2166   & 216.17          & 0.902          & 0.986 & 0.120          & 1441  & 274.00          & 0.857          & \textbf{0.980} & 0.172          & 4440 \\
(- Population Heuristic)       & 135.76          & 0.849          & 0.952          & 0.277          & 2166   & 208.73          & 0.897          & 0.986 & 0.114          & 1441  & 262.43          & 0.844          & \textbf{0.980} & 0.152 & 4445 \\
(- State-Level Context)        & 135.80          & 0.860          & 0.958          & 0.273          & 2166    & 209.57          & 0.897          & 0.986 & 0.116          & 1439 &            284.51          & 0.849          & 0.979          & 0.180          & 4437 \\
(GPT -> Gemini)                                & 129.21          & 0.859 & \textbf{0.987} & 0.279 & 2129   & 115.89         & 0.917          & 0.986          & 0.113          & 1310     & 250.58          & 0.864          & 0.979          & 0.161          & 3915 \\
(Gemini / 20->1 Candidate Entry)          & 132.28          & 0.807          & 0.981          & 0.305          & 2055   & 111.31         & 0.901        & 0.991 & 0.134          & 1251                  & 263.26          & 0.828          & 0.976          & 0.230          & 3601 \\
(Gemini / - Country Assisted Retrieval) & 131.45          & 0.848          & 0.980          & 0.286          & 2111          & 130.53         & \textbf{0.927} & 0.983          & 0.109          & 1307        & 349.80          & 0.860          & 0.960          & 0.174          & 3887 \\
(Gemini / - GeoNames Feature Types)     & 135.30          & 0.837          & 0.978          & 0.291          & 2110       & 96.86          & 0.904          & 0.990          & 0.126          & 1317  & 225.58 & 0.865          & \textbf{0.980} & 0.160          & 3910 \\
(Gemini / - Population Heuristic)       & 148.85          & 0.840          & 0.982          & 0.286          & 2130        & 108.27         & 0.914          & 0.989          & \textbf{0.104} & 1338    & 238.42          & 0.876 & 0.975          & \textbf{0.134} & 4301 \\
(Gemini / - State Level Context)        & 126.09 & 0.843          & 0.981          & 0.283          & 2126    & \textbf{92.37} & 0.920          & 0.991 & 0.113          & 1332 &       267.66          & 0.843          & 0.974         & 0.174          & 3917 \\ \hline
\end{tabular}
\caption{Comparison of RACCOON with baselines. Figures in bold are best values. Higher is better for Accuracy @161km (A@161) and Country Accuracy (CAcc). Lower is better for Mean Error (MErr) and AUC. Num indicates number of news articles that return a valid output. For ablation tests, - indicates that the component of RACCOON that has been removed, while -> indicates that a component has been replaced. Best values in every column are in boldface.}
\label{tab:final-results}
\end{table*}
We compare RACCOON with three baselines: (A) \textbf{Gazetteer Base Model}: In this case, spaCy NER\footnote{https://spacy.io/; Accessed on 5th December 2024.} is used to extract toponyms from news text. We extract tokens labeled as LOC, GPE or FAC. These toponyms are then geocoded using the Nominatim geocoder\footnote{https://nominatim.org/; Accessed on 5th December 2024.}, which is the geocoding service used by OpenStreetMap. Only exactly matching toponyms are geocoded. We take the first response from Nominatim as the matching gazetteer entry to infer coordinates and country for the toponym; (B) \textbf{LLM Base Model}: Our second baseline model uses an LLM to perform both toponym recognition and toponym resolution tasks. We prompt OpenAI's GPT-4o-mini LLM to extract location references from news article text and to infer coordinates and country for these references. Like with the gazetteer base model, we only compare geocoding results for extracted toponyms which are exact matches to those annotated in the evaluation dataset; (C) \textbf{RAG Base Model}: This is a simplified version of RACCOON using a retriever that does not implement Country-Assisted Retrieval, population re-ranking or additional context, and retrieves 10 candidate locations. 

\section{Results}
\begin{figure}
    \centering
    \includegraphics[width=\linewidth]{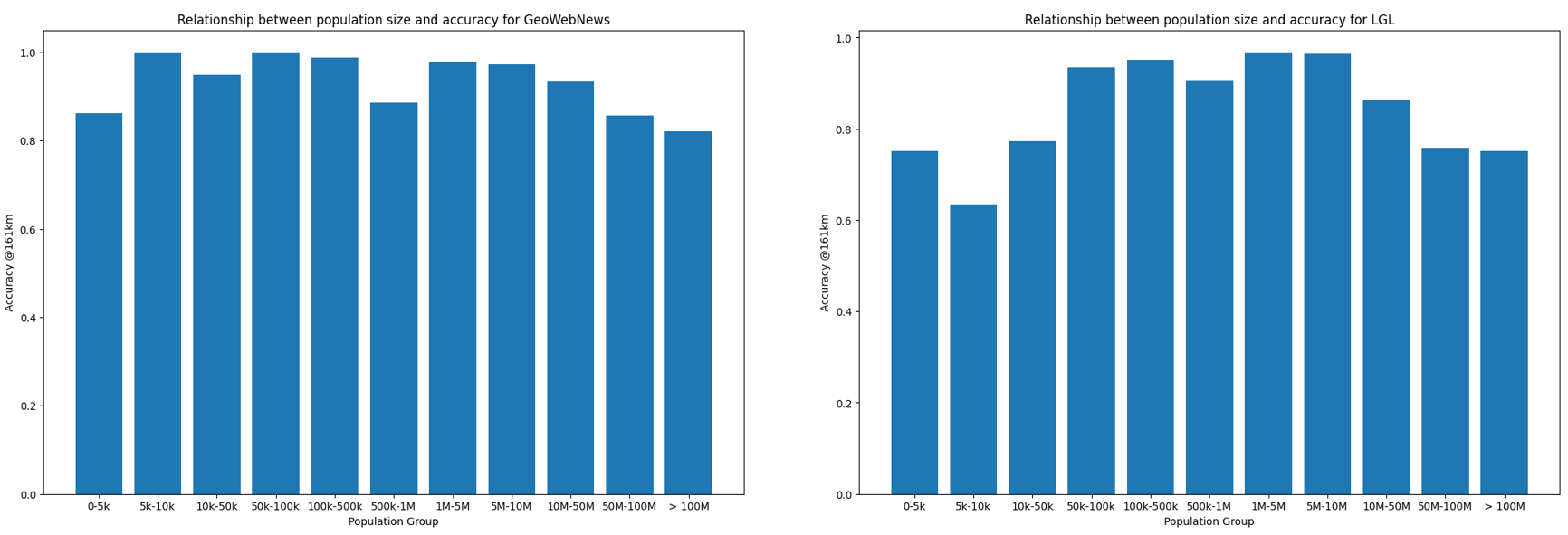}
    \caption{Relationship between population size and accuracy @161km for GeoWebNews and LGL datasets}
    \label{fig:population-bias}
\end{figure}

We now describe the performance of RACCOON in comparison with our aforementioned base models as well as a set of ablations in Table \ref{tab:final-results}. We point the reader to the caption of the table for a description of the labels. RACCOON performs the best for GeoVirus, particularly considering that it returns the most number of results (indicated by `Num'). RACCOON improves over its RAG Base counterpart across all metrics. In particular, large improvements in mean error for GeoWebNews and LGL indicate that the additional features have improved the RAG approach's accuracy. Furthermore, RACCOON consistently outperforms all baselines in AUC, showing the precision offered by linking direct geospatial information to an LLM compared to the LLM base model. Notably, RACCOON underperforms compared to the LLM Base, particularly in the mean error metric. 
A comprehensive discussion on better performance on ablation tests is left out due to lack of space, but are potential directions of research. However, we highlight one issue in particular: RACCOON underperforms on the LGL dataset compared to the ablation study conducted without using the population heuristic. As LGL is more biased towards local areas in the US compared to GeoVirus and GeoWebNews. This implies that RACCOON might suffer from population bias, where locations with smaller populations are geocoded with a lower accuracy compared to locations with larger populations.  To examine this possibility, we follow \citet{hu2024toponym} who graph the accuracy @161km of locations in different population buckets. RACCOON's results on the GeoVirus and LGL dataset, shown in Figure \ref{fig:population-bias} show lower accuracies at lower and higher population levels. However, since high-population locations have larger areas and are, therefore, more prone to gazetteer misalignment, this does suggest inherent population bias in RACCOON.

A manual analysis of errors also suggests areas for improvement. For example, toponyms containing common keywords such as ``North'' can be overlooked if they have a small population due to population bias. If the correct location is not included as a candidate locations, the model can give a completely incorrect answer while the LLM base model excels at placing locations in a generally correct area.

\section{Conclusion \& Future Work}
In this paper, we have presented RACCOON, a geocoding model that uses an RAG framework to help LLMs perform this important task. Our results showed that models based on LLMs have remarkable levels of accuracy across datasets. Furthermore, the usage of a RAG framework helps improve precision and consistency compared to a pure LLM base. However, we also revealed recall and population bias as two major shortcomings of RACCOON.

Our use of a statistical retriever was motivated by the costs of creating deep-learning embeddings over the extremely large GeoNames dataset. Exploring the capability of open-source LLMs could mitigate this issue and expand on our study of proprietary LLMs. Another potential area of interest is fine-tuning. While effective in connecting specific information to an LLM, RAG does not offer domain specific performance due to its reliance on pre-trained embeddings based on more common tasks such as text generation and NLP.
\bibliographystyle{ACM-Reference-Format}
\bibliography{sample-base}
\end{document}